\documentclass[letterpaper]{article} 
\usepackage{aaai25}  
\usepackage{times}  
\usepackage{helvet}  
\usepackage{courier}  
\usepackage[hyphens]{url}  
\usepackage{graphicx} 
\usepackage{natbib}  
\usepackage{caption} 
\usepackage{amsmath}
\usepackage{algorithm}
\usepackage{algorithmic}

\usepackage{newfloat}
\usepackage{listings}
\DeclareCaptionStyle{ruled}{labelfont=normalfont,labelsep=colon,strut=off} 
\floatstyle{ruled}
\newfloat{listing}{tb}{lst}{}
\floatname{listing}{Listing}
\title{Unsupervised Patch-GAN with Targeted Patch Ranking for Fine-Grained Novelty Detection in Medical Imaging}
\author{
    Jingkun Chen\textsuperscript{\rm 2 *},
     Guang Yang\textsuperscript{\rm 2}  \thanks{Co-first authors with equal contributions},Xiao Zhang\textsuperscript{\rm 5}, Jingchao Peng\textsuperscript{\rm 4}, Tianlu Zhang\textsuperscript{\rm 3}, Jianguo Zhang\textsuperscript{\rm 1,6,7} \thanks{Corresponding author: zhangjg@sustech.edu.cn, jungonghan77@gmail.com}, Jungong Han\textsuperscript{\rm 3 \dag} and Vicente Grau\textsuperscript{\rm 2}
}
\affiliations{
    \textsuperscript{\rm 1}Department of Computer Science and Engineering, the Southern University of Science and Technology, Shenzhen, China.\\

    \textsuperscript{\rm 2}Department of Engineer Science, University of Oxford, Oxford, UK\\
    \textsuperscript{\rm 3}Department of Automation, Tsinghua University, Beijing, China\\
    \textsuperscript{\rm 4}Warwick Manufacturing Group, University of Warwick, Coventry, UK\\
    \textsuperscript{\rm 5}School of Information Science and Technology, Northwest University, Xi'an, China\\
    \textsuperscript{\rm 6}Peng Cheng Lab, Shenzhen, China\\
    \textsuperscript{\rm 7}Guangdong Provincial Key Laboratory of Brain-inspired Intelligent Computation, Shenzhen, China\\

}

\usepackage{bibentry}

\begin{document}

\maketitle

\begin{abstract}
Detecting novel anomalies in medical imaging is challenging due to the limited availability of labeled data for rare abnormalities, which often display high variability and subtlety. This challenge is further compounded when small abnormal regions are embedded within larger normal areas, as whole-image predictions frequently overlook these subtle deviations. To address these issues, we propose an unsupervised Patch-GAN framework designed to detect and localize anomalies by capturing both local detail and global structure. Our framework first reconstructs masked images to learn fine-grained, normal-specific features, allowing for enhanced sensitivity to minor deviations from normality. By dividing these reconstructed images into patches and assessing the authenticity of each patch, our approach identifies anomalies at a more granular level, overcoming the limitations of whole-image evaluation. Additionally, a patch-ranking mechanism prioritizes regions with higher abnormal scores, reinforcing the alignment between local patch discrepancies and the global image context. Experimental results on the ISIC 2016 skin lesion and BraTS 2019 brain tumor datasets validate our framework’s effectiveness, achieving AUCs of 95.79\% and 96.05\%, respectively, and outperforming three state-of-the-art baselines.
\end{abstract}

\section{Introduction}
Novelty detection in medical imaging is essential for identifying unseen abnormalities that are not represented in limited labeled datasets, a task critical for early diagnosis and effective treatment planning. In medical practice, manually annotating the extensive variety of possible abnormalities—often irregular in shape, size, and texture—is infeasible, particularly in domains such as dermatology and neuroimaging \cite{yang2025contrast, chen2022semi,duan2024wearable}. These abnormalities may be subtle and diverse, requiring unsupervised methods capable of leveraging predominantly normal data to reliably detect new patterns, enabling accurate anomaly identification without exhaustive manual labeling \cite{luo2024lesion, chen2020deep}.

Despite its importance, medical novelty detection faces two fundamental challenges. Small abnormalities within large normal regions are often overlooked in global predictions, as they resemble normal variations \cite{gao2023video}. This challenge is compounded by current methods' inability to seamlessly integrate localized detection with global context, which is crucial for accurately identifying these small anomalies in predominantly normal images. Approaches that either emphasize global features or treat patches in isolation often fail to capture the nuanced relationship between local anomalies and the overall image, limiting their effectiveness in medical novelty detection \cite{chen2024dynamic}.

To address these challenges, we propose an innovative unsupervised Patch-GAN framework for medical novelty detection that combines local and global perspectives to enhance both sensitivity and localization accuracy. Our approach reconstructs masked normal images, allowing the model to learn intricate local details representative of the normal class—such as texture, structure, and intensity—without any explicit labels. By partitioning the reconstructed images into patches and evaluating each patch independently, our model identifies fine-grained discrepancies that may indicate anomalies. This patch-level analysis increases the model’s sensitivity to small, isolated abnormalities while simultaneously establishing a coherent link between local patches and the overall image context, a key aspect for accurate anomaly localization.

To further strengthen the model’s ability to detect and prioritize abnormal regions, we introduce a novel patch-ranking mechanism. This mechanism ranks patches by their abnormal scores, selectively highlighting areas with the highest likelihood of abnormality and integrating them within the broader image context. By correlating these high-anomaly patches with the global structure of the image, our approach effectively balances detailed local detection with contextual awareness, enabling the model to not only detect but also accurately localize small, subtle anomalies. This innovative strategy addresses the limitations of existing methods by providing a multi-scale analysis that adapts to the complexity of medical imaging data.

We validate our framework on two public medical imaging datasets—ISIC 2016 for skin lesions and BraTS 2019 for brain tumors—demonstrating that our method consistently outperforms state-of-the-art approaches in sensitivity and specificity. Our contributions are fourfold:

\begin{itemize}
    \item We introduce an unsupervised Patch-GAN framework that reconstructs masked images to learn comprehensive normal patterns, enabling precise novelty detection at the local level.
    \item We propose a patch-based detection approach that enhances the model’s sensitivity to small abnormalities while maintaining a coherent local-global relationship within each image.
    \item We develop a patch-ranking mechanism that prioritizes high-anomaly patches, improving both detection accuracy and the model’s ability to integrate patch-level insights with the overall image structure.
    \item Experimental results demonstrate the effectiveness of our approach, achieving state-of-the-art performance on multiple datasets.
\end{itemize}

\section{Related Work} 

Novelty detection in medical imaging has evolved from classical unsupervised algorithms to sophisticated deep learning models, each attempting to capture the subtle and often complex anomalies characteristic of medical data. Here, we review these advancements and highlight the distinctions and innovations of our unsupervised Patch-GAN framework.

Early methods in novelty detection focused on representing the normal class through a dense, low-dimensional distribution, with anomalies identified as deviations from this normal range. Principal Component Analysis (PCA) \cite{shyu2003novel} was among the foundational approaches, projecting data into a low-dimensional space to isolate outliers. Extending this, Soft PCA \cite{aggarwal2017introduction} introduced the Mahalanobis distance to quantify deviations from the normal class center, improving separation of normal and abnormal samples. Other approaches, such as one-class Support Vector Machines (SVM) \cite{perdisci2006using} and local density-based methods like Local Outlier Factor (LOF) \cite{breunig2000lof} and Local Correlation Integral (LOCI) \cite{papadimitriou2003loci}, used decision boundaries or local density to identify anomalies. Isolation Forest (iForest) \cite{liu2008isolation} offered a different strategy, using randomly partitioned hyperplanes to isolate outlying samples. While these classical methods provide a baseline for novelty detection, their limited capacity to model the complex textures and structural nuances in medical images often leads to reduced performance when detecting subtle or small-scale abnormalities.

With the advent of deep learning, reconstruction-based models, particularly those using GANs, have shown promise in capturing richer representations of normal data. AnoGAN \cite{schlegl2017unsupervised} was one of the first to leverage GANs for medical novelty detection, reconstructing normal-class images and identifying anomalies through reconstruction error. However, this pixel-level adversarial learning approach proved computationally expensive, as it required a costly remapping of each sample in the latent space. To address this inefficiency, Efficient-GAN-abnormal \cite{zenati2018efficient} optimized the distribution mapping, though the computational burden remained significant. Schlegl et al. \cite{schlegl2019f} further improved speed by incorporating a pretrained encoder, thereby bypassing the need for remapping, but at the cost of added model complexity.

Further advancements in GAN-based models, such as Ganomaly \cite{akcay2018ganomaly} and OCGAN \cite{perera2019ocgan}, explored the use of compact latent spaces to distinguish normal from abnormal samples. These methods assume that latent spaces trained exclusively on normal data will form a compact distribution, enabling easier identification of anomalies as deviations. However, the high generalization capability of deep models can blur these latent representations, especially for subtle or small abnormalities, limiting detection precision \cite{shen2020counterfeit}. To mitigate this, Deng et al. \cite{deng2022anomaly} proposed a reverse distillation framework, using a teacher-student architecture to align multi-scale features and improve sensitivity to anomalies. Despite its improved performance, this approach adds computational overhead due to the complexity of multi-scale feature alignment.

Beyond GANs, novel paradigms have emerged to further refine anomaly detection. Diffusion models \cite{wolleb2022diffusion}, as unsupervised generative methods, exploit iterative denoising to capture intricate data distributions, demonstrating strong potential for novelty detection. Similarly, self-supervised approaches have utilized translation-consistent features, combining image- and feature-level reconstructions to improve robustness \cite{zhao2021anomaly, pinaya2022unsupervised}. In addition, VQ-VAE coupled with transformer-based pretraining \cite{tian2023self} has introduced hierarchical feature extraction mechanisms, enhancing latent space quality. Complementing these, uncertainty refinement strategies \cite{cai2023dual} have provided robust anomaly predictions by integrating uncertainty modeling into the detection pipeline.

Motivated by these advancements, we present a novel unsupervised Patch-GAN framework designed to overcome the limitations of global reconstruction methods. Instead of reconstructing entire images—which can obscure small or localized anomalies—our method focuses on masked image reconstructions at the patch level. By dividing reconstructed images into patches and analyzing them independently, our framework ensures that fine-grained abnormalities are effectively captured. To further enhance performance, we propose a patch-ranking mechanism that prioritizes patches with high anomaly scores, enabling the integration of localized anomaly signals into a cohesive global context. This patch-wise approach not only improves sensitivity to small abnormalities but also provides a scalable, computationally efficient solution for medical imaging tasks, pushing the boundaries of current novelty detection frameworks.

\section{Method}

\begin{figure}[ht]
  \centering
  \includegraphics[height=4.5cm,width=8.2cm]{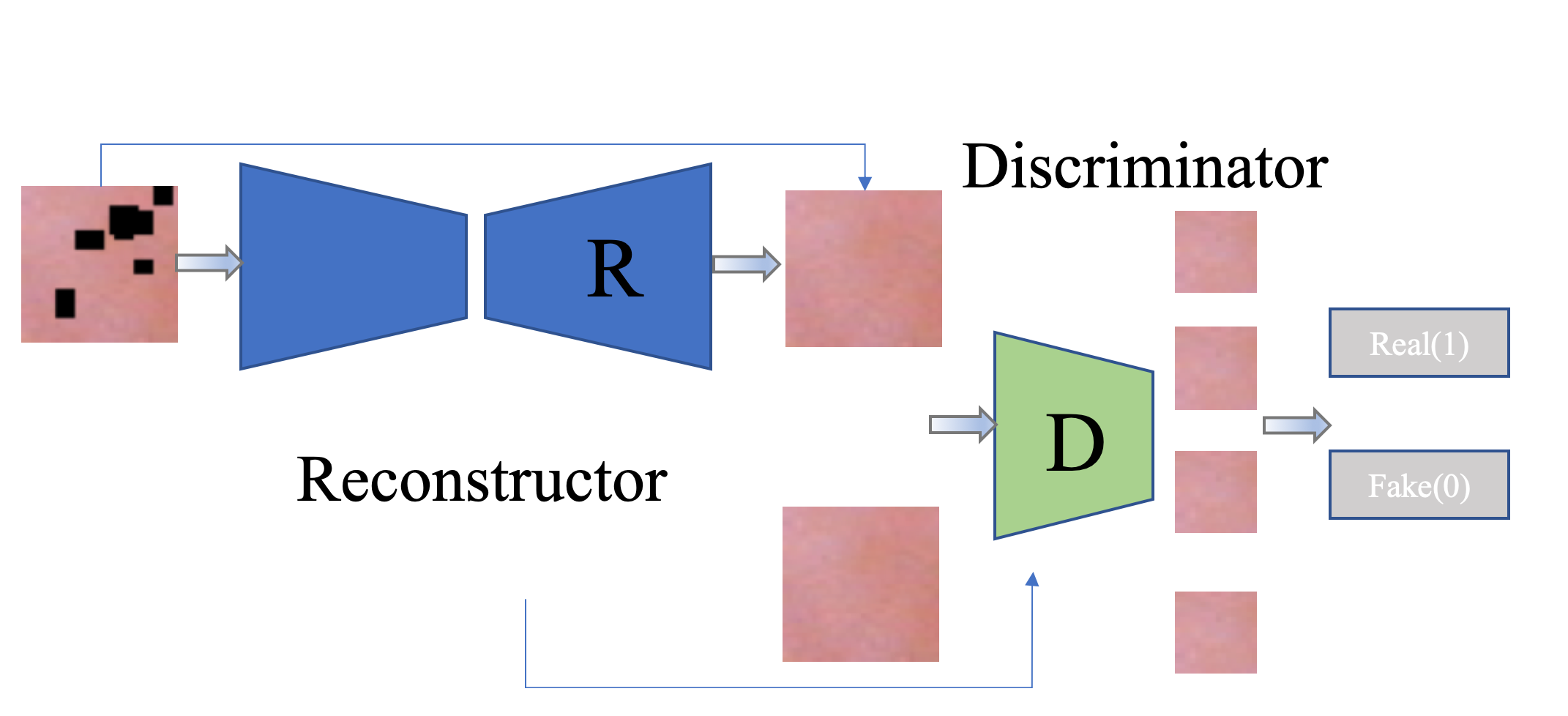}
  \caption{Overview of the proposed one-class classification framework, illustrating the key components for novelty detection, including patch-level reconstruction and targeted patch ranking.}
  \label{alocc_model}
\end{figure}

Our proposed unsupervised Patch-GAN framework (Fig. \ref{alocc_model}) is designed to capture both local and global image features for effective novelty detection in medical images. The method integrates two primary components: a mask reconstructor, which restores pixel-level details from normal images to capture fine-grained characteristics, and a patch-ranking strategy that enhances sensitivity to localized abnormalities. This combination enables precise novelty detection by leveraging patch-based evaluations and focused ranking of high-discrepancy regions.

\begin{figure}[ht]
  \centering
   \includegraphics[height=4cm,width=4cm]{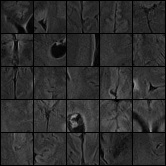}
   \includegraphics[height=4cm,width=4cm]{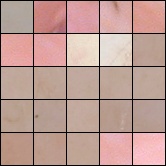}
   
  \includegraphics[height=4cm,width=4cm]{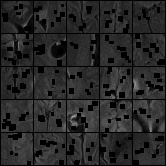}
  \includegraphics[height=4cm,width=4cm]{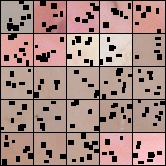}
  \caption{Examples of training images used in the model: (top row) original normal images from the brain (left) and skin (right) datasets; (bottom row) corresponding masked images used for self-supervised learning, where black boxes denote masked regions.}
  \label{noise_image}
\end{figure}

\noindent\textbf{Mask Reconstructor}: To model the detailed structures of normal images, we employ a mask reconstructor based on an encoder-decoder architecture. This module reconstructs missing regions in masked normal images (Fig. \ref{noise_image}), learning to represent the intrinsic texture, color, and structural attributes unique to the normal class. During training, masked images are passed through the reconstructor, which learns to complete the missing areas, reinforcing the model’s understanding of pixel-level normalcy. The reconstruction loss encourages accurate restoration of normal regions while exposing abnormalities in areas that deviate. This loss function is given by:

\begin{equation}
    \mathcal{L}_{R(x)} = \left\| x - R(z) \right\|^{2} ,
    \label{reconstructor}
\end{equation}
where $x$ is the original image, $z$ is the masked image input, and $R(z)$ represents the reconstructed image generated by the reconstructor. This loss encourages the model to accurately reconstruct the entire image, learning normal patterns without any anomalies.

\noindent\textbf{Patch Discriminators}: To enhance the model’s sensitivity to small-scale anomalies, we employ a PatchGAN discriminator that evaluates images on a patch-wise basis. Rather than outputting a single score for the entire image, the PatchGAN discriminator generates a grid of scores, with each score reflecting the discriminator’s confidence that a particular patch in the reconstructed image resembles a corresponding patch in the original image. This patch-based assessment enhances the model’s ability to focus on localized regions, making it particularly effective for detecting small, subtle abnormalities that may otherwise be overlooked in global assessments.

The PatchGAN discriminator operates by sliding a receptive field over the reconstructed image, comparing corresponding patches from the original and reconstructed images. The discriminator loss is calculated as follows:

\begin{equation}
    \mathcal{L}_{D(x)} = \sum_{i=1}^{N}  \left\| x_i - \tilde{x}_i \right\|^{2} ,
    \label{discriminator}
\end{equation}
where $x_i$ and $\tilde{x}_i$ are corresponding patches from the original image $x$ and the reconstructed image $R(z)$, respectively, and $N$ is the number of patches. This patch-based comparison helps the discriminator identify local discrepancies, allowing for precise localization of subtle anomalies within the image.

This approach allows the model to capture and highlight local discrepancies directly between patches, avoiding the need for multiple independent discriminators and thus balancing efficiency with localized sensitivity.

\noindent\textbf{Adversarial Training of Reconstructor and Discriminator}:
The adversarial training of the mask reconstructor and PatchGAN discriminator forms a min-max optimization framework. In this setting, the reconstructor attempts to produce realistic patches that mimic the original image, while the PatchGAN discriminator distinguishes between real and reconstructed patches. This adversarial setup combines the objectives of reconstruction accuracy and patch-level realism, refining the model’s capability to capture normal patterns and highlight abnormal ones. The adversarial objective is defined as:

\begin{equation}
\begin{aligned}
R_{Min} D_{Max} V(D,R) = R_{x\sim P_{x}}\{log~[D(x)]\} \\ +R_{z\sim P_{z}}\{log~(1-[D(R(z)])\} ,
	\label{adversarial}
\end{aligned}
\end{equation}
where $P_{x}$ and $P_{z}$ denote the distributions of the original and masked images, respectively. This setup ensures that the reconstructor closely approximates normal patches, while abnormal patches become distinct and easily identifiable.

The total adversarial loss is given by:

\begin{equation}
	\mathcal{L}_{adv} =\lambda \mathcal{L}_{R} +  \mathcal{L}_{D},
	\label{total_loss}
\end{equation}
where $\lambda$ is a weight balancing the reconstruction and adversarial loss components, empirically set to 0.2 for optimal results.

\noindent\textbf{Patch Ranking Strategy}: To further enhance localization, we introduce a patch-ranking strategy that prioritizes patches based on their reconstruction loss. This strategy assigns a higher weight to patches with greater discrepancies, effectively ranking patches by abnormal scores. By emphasizing high-discrepancy regions, this approach enables the model to capture subtle abnormalities and localize them within the broader image context.

The patch-ranking strategy calculates an overall abnormal score by summing weighted reconstruction errors for patches, focusing on those with the most significant discrepancies. The abnormal score is defined as:

\begin{equation}
	 \mathcal{L}_{score} = \sum_{i=1}^{N}\lambda_{i} \mathcal{L}_{R\_{patch\_i}},
	\label{abnormal_score}
\end{equation}
where $\mathcal{L}_{R\_{patch\_i}}$ represents the reconstruction loss for each patch and $\lambda_{i}$ is a weight prioritizing patches with high abnormal scores. In our experiments, we assigned higher weights (e.g., 1.1 for the most anomalous patch and 1.0 for the next) to the patches with the greatest reconstruction discrepancies. This weighting scheme enhanced the model’s sensitivity to critical abnormal regions, ensuring that these areas received greater emphasis during anomaly scoring.

By combining masked image reconstruction, patch-wise PatchGAN discrimination, and a targeted patch-ranking mechanism, our framework achieves a novel balance between local sensitivity and global image coherence. This unsupervised approach provides an effective solution for identifying and localizing complex anomalies in medical imaging.

\section{Experiments}
To assess the effectiveness of our unsupervised Patch-GAN framework for novelty detection, we conduct experiments on two widely recognized medical imaging datasets: ISIC 2016 for skin lesion analysis and BraTS 2019 for brain tumor detection. We compare our method with three state-of-the-art novelty detection techniques, including Autoencoder (AE), One-Class SVM (OCSVM), and Adversarially Learned One-Class Classification (ALOCC). We also evaluate our framework with and without the patch-ranking strategy to quantify the impact of prioritized anomaly scoring.
\begin{figure*}[ht]
  \centering
  \includegraphics[height=6cm,width=17.3cm]{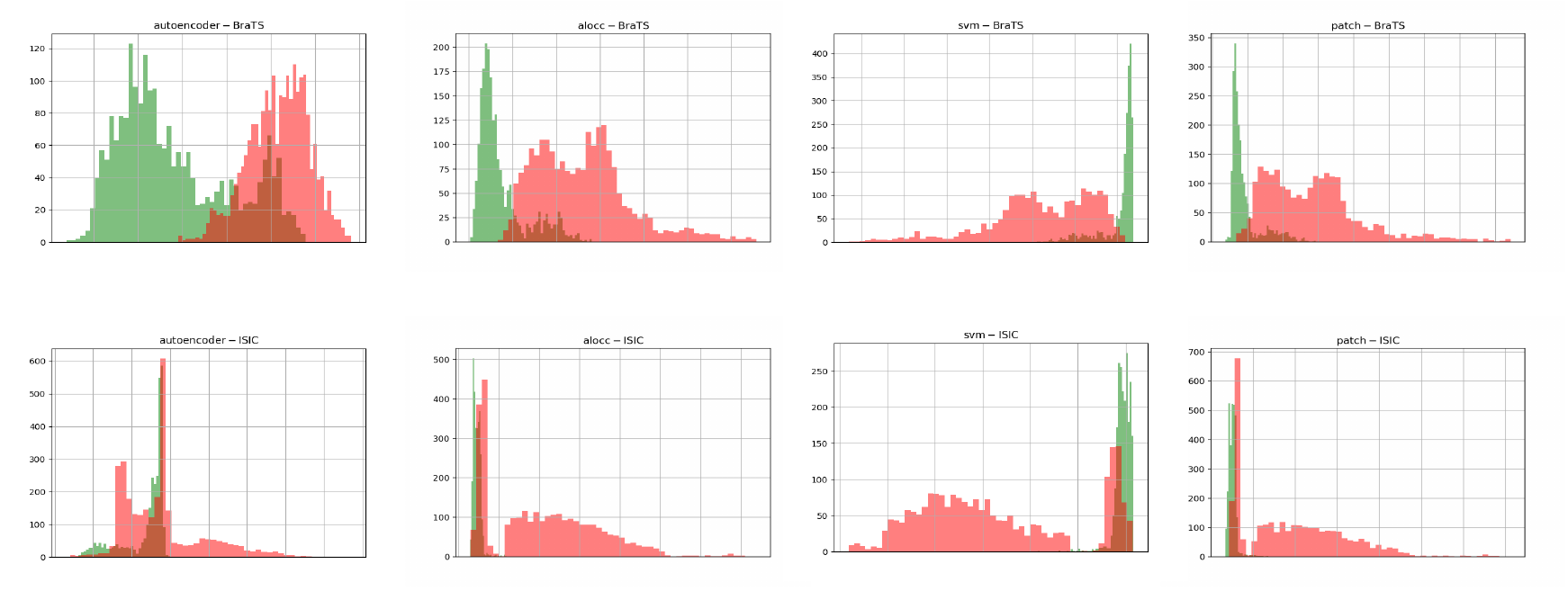}
  \caption{Distribution of abnormal scores for AE, ALOCC, OCSVM, and the proposed method on the BraTS 2019 and ISIC 2016 datasets (left to right), illustrating the separation between normal and abnormal classes across methods.}
  \label{distribution}
\end{figure*}
\subsection{Datasets}
\begin{figure}[H]
  \centering
  \includegraphics[height=4cm,width=4cm]{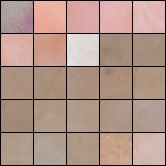}
  \includegraphics[height=4cm,width=4cm]{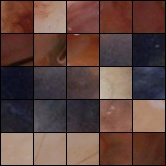}

  \caption{ISIC 2016 samples: normal (left) and abnormal (right).}
  \label{skin_samples}
\end{figure}

\noindent\textbf{ISIC 2016} \cite{gutman2016skin}: The ISIC 2016 dataset (Fig. \ref{skin_samples}) provides high-resolution images of benign and malignant skin lesions, offering a relevant context for evaluating subtle visual abnormalities. We partition this dataset into a training set of 34,104 normal samples and a test set comprising 3,000 normal and 3,000 abnormal samples. Each image is divided into 32×32 patches, enabling the model to focus on fine-grained local structures and effectively detect small irregularities. Representative samples for each class are shown in Fig. \ref{skin_samples}, illustrating the diverse visual characteristics of normal and abnormal skin.

\begin{figure}[H]
  \centering
  \includegraphics[height=4cm,width=4cm]{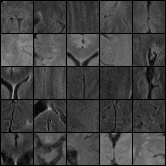}
  \includegraphics[height=4cm,width=4cm]{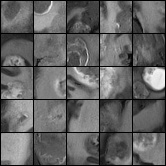}

  \caption{BraTS 2019 samples: normal (left) and abnormal (right).}
  \label{brain_samples}
\end{figure}

\noindent\textbf{BraTS 2019} \cite{menze2014multimodal}: The BraTS dataset (Fig. \ref{brain_samples}) consists of multimodal MRI scans of brain tumors, with only the FLAIR modality used in this study to target tumor-related anomalies. This dataset is partitioned into 36,073 normal samples for training and a test set of 2,200 normal and 2,200 abnormal samples. Each FLAIR image is processed similarly to ISIC, with patching used to enhance sensitivity to localized abnormalities, such as tumor regions. Sample images from each class are displayed in Fig. \ref{brain_samples}.

\subsection{Evaluation Metrics}

To quantitatively evaluate our model’s performance, we use the Area Under the Receiver Operating Characteristic Curve (AUC), a widely recognized metric for binary classification tasks in medical imaging. The AUC provides a robust assessment of sensitivity and specificity across different decision thresholds, making it particularly suitable for comparing novelty detection performance across multiple methods.

\subsection{Performance Comparison}
The AUC results for all methods on the ISIC and BraTS datasets are shown in Table \ref{comparsion_results}. Our Patch-GAN framework, equipped with the patch-ranking strategy, achieves superior performance, with AUCs of 95.79\% on ISIC and 96.05\% on BraTS, outperforming baseline methods such as ALOCC, OCSVM, and Autoencoder. Specifically, our method shows a 0.9\% improvement over ALOCC on the ISIC dataset and a 1.0\% increase on BraTS, highlighting the effectiveness of our framework in distinguishing abnormal patterns that are visually subtle. Notably, the improvement over AE (36.7\% on ISIC and 3.4\% on BraTS) indicates that reconstructing patches in the context of a local-global framework enhances novelty detection capabilities substantially.

\begin{figure}[ht]
  \centering
  \includegraphics[height=5cm,width=5.5cm]{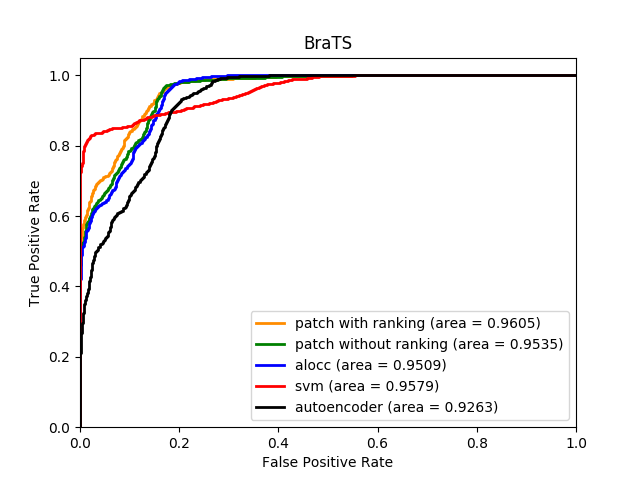}
  \includegraphics[height=5cm,width=5.5cm]{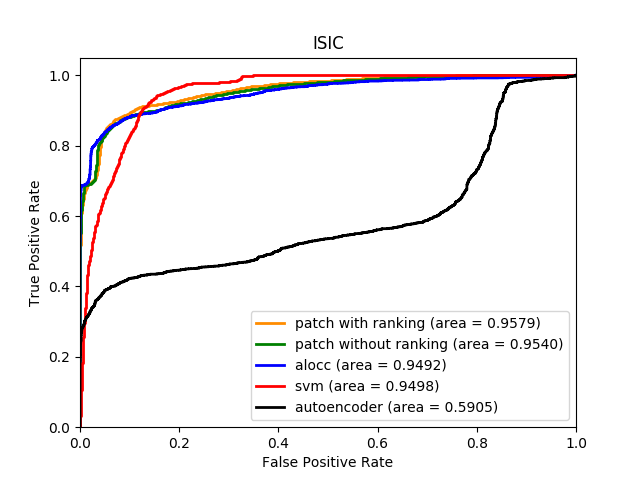}

  \caption{ROC curves for five novelty detection methods on the BraTS 2019 (left) and ISIC 2016 (right) datasets, illustrating comparative performance across models.}
  \label{roc}
\end{figure}
Fig. \ref{roc} illustrates this improvement by displaying ROC curves for our method and the baseline models on both datasets, where our approach shows a sharper distinction between normal and abnormal samples.

\subsection{Effect of Patch-Ranking Strategy}
The addition of our patch-ranking strategy significantly improves novelty detection, as shown in Table \ref{comparsion_results_ours}. Without ranking, our Patch-GAN achieves AUCs of 95.40\% on ISIC and 95.35\% on BraTS. With patch-ranking enabled, these scores rise to 95.79\% and 96.05\%, respectively, suggesting that prioritizing high-discrepancy patches enables the model to localize anomalies more effectively within complex normal structures.

\begin{table}[ht]
\renewcommand{\arraystretch}{1.3}
\caption{AUC comparison on ISIC and BraTS datasets, showing the effectiveness of the proposed Patch-GAN with patch-ranking.}
\label{comparsion_results}
\centering
\small 
\begin{tabular}{|c||c||c|}
\hline
           & ISIC       & BraTS \\ \hline

        AE      &   59.05\%          &    92.63\%   \\

        OCSVM \cite{perdisci2006using}     &   94.98\%          &    95.79\%   \\

        ALOCC \cite{sabokrou2018adversarially}               &   94.92\%          &    95.09\%   \\

        \textbf{Ours (Patch with ranking)}                &   \textbf{95.79\%}          &  \textbf{96.05\%}     \\ 
        
\hline
\end{tabular}
\end{table}

\begin{table}[ht]

\renewcommand{\arraystretch}{1.3}
\caption{AUC comparison on ISIC and BraTS datasets with and without patch-ranking.}
\label{comparsion_results_ours}
\centering
\begin{tabular}{|c||c||c|}
\hline
		  			& ISIC  		& BraTS \\ \hline
		Patch without ranking    				&   95.40\%			&  95.35\%     \\

        Patch with ranking    				&   \textbf{95.79\%}			&  \textbf{96.05\%}     \\ 
		
\hline
\end{tabular}
\end{table}

\subsection{Score Distribution Analysis}
To further investigate our method’s discriminative capability, we analyze the distribution of abnormal scores across methods on both datasets, as shown in Fig. \ref{distribution}. The distributions demonstrate that our method, particularly with patch-ranking enabled, yields a well-separated and narrower distribution of scores between normal and abnormal classes, indicating its capacity to isolate subtle abnormalities with minimal overlap. In contrast, the score distributions of baseline models, such as AE and ALOCC, show substantial overlap between normal and abnormal classes, which can lead to less accurate novelty detection. This suggests that our model not only achieves high overall performance but also excels in producing reliable, interpretable scores that support clinical decision-making.

The experimental results substantiate the efficacy of our unsupervised Patch-GAN framework, particularly when augmented with the patch-ranking strategy. By reconstructing masked images at a patch level and prioritizing regions with higher abnormal scores, our model effectively bridges local and global image features, leading to precise localization of small and subtle abnormalities. This combination of patch-level assessment and focused ranking significantly outperforms state-of-the-art methods on both ISIC and BraTS datasets. Our findings validate that the proposed framework not only enhances novelty detection performance but also provides interpretable insights that align well with clinical requirements, marking a significant advancement in medical novelty detection.

\section{Conclusion}
We introduced an unsupervised Patch-GAN framework for medical novelty detection, designed to enhance localization of subtle abnormalities through patch-level reconstruction and targeted patch ranking. Our approach, leveraging both local detail and global context, achieved competitive AUCs of 95.79\% on ISIC 2016 and 96.05\% on BraTS 2019, outperforming existing methods. While GAN-based models proved effective, future work may explore diffusion models or other advanced generative frameworks to further increase sensitivity to fine-grained anomalies, advancing the precision of novelty detection in complex medical imaging tasks.
\section{Acknowledgments}
This work is supported in part by National Key Research and Development Program of China (2021YFF1200800) and National Natural Science Foundation of China (Grant No. 62276121 and 12326604).

\end{document}